\documentclass{article} 
\usepackage{hyperref}
\usepackage{url}

\usepackage{iclr2026_conference,times}

\usepackage{amsmath,amsfonts,bm}

\def\eqref#1{equation~\ref{#1}}

\def\1{\bm{1}}

\DeclareMathAlphabet{\mathsfit}{\encodingdefault}{\sfdefault}{m}{sl}
\SetMathAlphabet{\mathsfit}{bold}{\encodingdefault}{\sfdefault}{bx}{n}

\usepackage{graphicx}
\usepackage{wrapfig}
\usepackage{amssymb}
\usepackage{caption}
\usepackage{subcaption}
\usepackage{algpseudocode}
\usepackage[ruled,vlined]{algorithm2e}
\usepackage{booktabs}
\usepackage{multirow}

\title{Pitfalls of Unlabeled Disagreement-Based Drift Detection in Streaming Tree Ensembles}

\author{Lara Sá Neves, Afonso Lourenço, Goreti Marreiros\\
GECAD, ISEP, Polytechnic of Porto, Portugal \\
\texttt{\{lspsn,fonso,mgt\}@isep.ipp.pt} \\
\And
Lizy K. John \\
The University of Texas at Austin, USA \\
\texttt{ljohn@ece.utexas.edu} \\
}

\iclrfinalcopy 
\begin{document}

\maketitle

\begin{abstract}
Detecting concept drift in high-speed data streams remains challenging, particularly when models must operate on unlabeled data and avoid false alarms caused by benign shifts. While disagreement-based uncertainty has shown promise in neural networks, its adaptation to ensembles of incremental decision trees (IDTs) remains largely unexplored. We investigate this approach by constructing batch-specific disagreement measures via label flipping in ensemble members and evaluating their effectiveness for drift detection in tabular data streams. Our experiments show that, although this method performs well in ensembles of multi-layer perceptrons (MLPs), it consistently underperforms loss-based detectors when applied to IDTs. We attribute this behavior to the intrinsic rigidity of IDTs: learning primarily through structural expansion, with limited parameter adaptation, restricts model plasticity and prevents disagreement from reliably reflecting learning potential. Recent work on restructuring IDTs using their intrinsic decomposition into non-overlapping rules offers a promising direction for improving adaptability.
\end{abstract}

\section{Introduction}

Handling change in high-speed data streams is challenging due to heavy concept drifts. Effective monitoring algorithms should (R1) operate on unlabeled deployment data to detect model deterioration and (R2) resist non-deteriorating shifts with few samples. While existing data-based drift detectors perform well on unlabeled data (R1) \citep{Xuan2021,Wan2021}, they often generate false positives when shifts are benign (R2). Many methods track changes in classifier posterior distributions \citep{lindstrom2013drift,lughofer2016recognizing,lu2025early}, which can indicate uncertainty. However, such estimates may be unreliable when models continuously adapt to evolving streams. To address this, we propose batch‑specific uncertainty, which is more practical than prequential metrics. Streaming models should focus on reliability in the current distribution rather than hypothetical generalization. Similarly to transductive reasoning, measuring how conflicting information in a batch affects the model rather than relying on accumulated past uncertainty.

A prominent example is the model disagreement framework \citep{yu2019unsupervised,jiang2021assessing,rosenfeld2023almost,ginsberg2022learning}. To date, it has been studied mainly with expressive neural networks trained in large batches, which struggle on tabular streams due to slow convergence, overwritten weights, and limited inductive advantage \citep{sahoo2017online}. For tabular data, ensembles of incremental decision trees (IDTs) remain state-of-the-art, leveraging fast online convergence and tree replacement via loss-based drift detectors \citep{adwin,gama2004learning}, with extensions incorporating unlabeled data through self-training, unsupervised drift detection, and active learning \citep{gomes2025sleade}. This raises a key question: can the disagreement framework be adapted for tree-based streaming ensembles? To implement this, we exploit the fact that in binary classification, arbitrarily flipping labels for each ensemble component can create diverse, disagreeing representations, a simple yet effective way to design a true disagreeing critic \citep{rosenfeld2023almost,ginsberg2022learning,pagliardini2022agree,chuang2020estimating}. Surprisingly, we find this strategy performs poorly across nearly all evaluated streams for ensembles of IDTs, but not multi-layer perceptrons (MLPs). We hypothesize that disagreement among IDTs fails to provide reliable signals of concept change, not due to flaws in the detection logic, but because the underlying learners lack the plasticity needed for a disagreement critic to capture their learning potential.

\section{Theory}

For a stream of drifting concepts $D_i$, learners update $\theta_t$ incrementally to minimize risk on $D_t$:
\begin{equation}
    \theta_t := \text{Alg}_t(\theta_{t-1}, \mathcal{L}_t), \mathcal{L}_t = \sum_{i=1}^t \mathbb{E}_{(x,y)\sim D_i}[\ell(y,h_{\theta_{t-1}}(x))].
\end{equation}
Storing all past data is impractical. Tree-based learners address this via approximations, e.g., incremental information gain with Hoeffding bounds, expanding only when differences between best and second-best splits are significant \citep{domingos2000mining}. This operation can be described as:

\textbf{Lemma 1 (Incremental labeled update).} For $h \in \mathcal{H}$ and history model $h_{\theta_{t-1}}$:
\begin{equation}
\varepsilon_{D_t}(h) = \varepsilon_{D_t}(h, h_{\theta_{t-1}}) + \varepsilon_{D_t}(h_{\theta_{t-1}}),
\end{equation}
where $\varepsilon_{D_t}(h, h_{\theta_{t-1}})$ denotes the one-hot disagreement. However, this bound is insufficient under drifting unlabeled distributions. Since manual labeling is infeasible in true streams, meaningful error bounds need a notion of distributional distance, e.g. $\mathcal{H}\Delta\mathcal{H}$-divergence \citep{kifer2004detecting}.

\textbf{Lemma 2 (Drift-based update).}  Assuming a binary hypothesis class capable of discriminating $\mathcal{D}_{t-1}$ and $\mathcal{D}_t$ \citep{ben2010theory}, i.e., whose
$\mathcal{H}\Delta\mathcal{H}$ class contains all pairwise exclusive-ors:
\begin{equation}
\varepsilon_{D_t}(h) \le \varepsilon_{D_t}(h, h_{\theta_{t-1}}) + \varepsilon_{D_{t-1}}(h_{\theta_{t-1}}) + \frac{1}{2}\Delta(h_{\theta_{t-1}}),
\end{equation}

\begin{wrapfigure}{r}{0.6\textwidth}  
    \centering
    \vspace{-0.7cm}
    \includegraphics[width=0.6\textwidth]{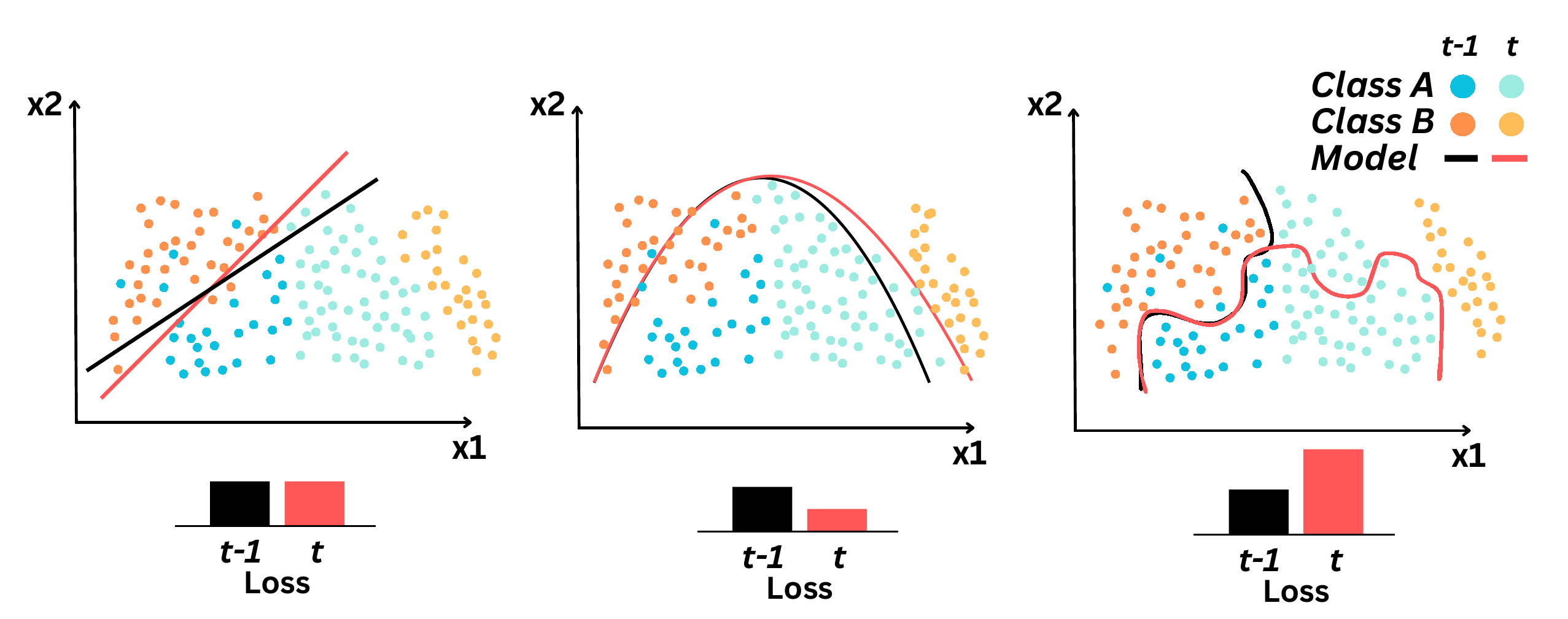}
    \vspace{-0.7cm}
    \caption{Drift detection across complexities: (left) loss-based false negative on over-regularized model, (center) data-based false positive for a true matching model complexity, (right) both successful in overly complex model.}
    \label{fig:1}
    \vspace{-0.7cm}
\end{wrapfigure}

While vacuous in practice, this bound suggests that (1) the conservative splitting and parent hyper-rectangles of $h_{\theta_{t-1}}$ act as a regularizer for $h$; (2) bias is minimized only if $h_{\theta_{t-1}}$ is localized around $\mathcal{D}_t$; and (3) useful drift detectors must account for both data and model complexity (Fig.~\ref{fig:1}): if $\mathcal{D}_{t-1}$/$\mathcal{D}_t$ are similar, the bound is small and $h_{\theta_{t-1}}$ can be reused; otherwise, $h$ is updated via pruning, regrowing, or ensemble modification.

This motivates bounding error relative to the previous model rather than the entire hypothesis class, as the true labeling function $y^*$ and drifted distribution $\mathcal{D}t$ are not adversarial. Hence, detection can exploit $\Delta(h_{\theta_{t-1}})$ with alternative hypotheses to obtain more practical bounds under drift:

\begin{minipage}[t]{\textwidth} 
\textbf{Lemma 3 (Disagreement-based update).} Let $h^* = \arg\max_{h' \in \mathcal{H}'} \Delta(h_{\theta_{t-1}},h')$, $\mathcal{H}'$ per $h_{\theta_{t-1}}$:
\begin{equation}
\varepsilon_{D_t}(h) \le \varepsilon_{D_t}(h,h_{\theta_{t-1}}) + \varepsilon_{D_{t-1}}(h_{\theta_{t-1}}) + \frac{1}{2}\Delta(h_{\theta_{t-1}},h^*),
\end{equation}
\end{minipage}

\begin{wrapfigure}{r}{0.6\textwidth}  
    \centering
    \vspace{-0.5cm}
    \includegraphics[width=0.6\textwidth]{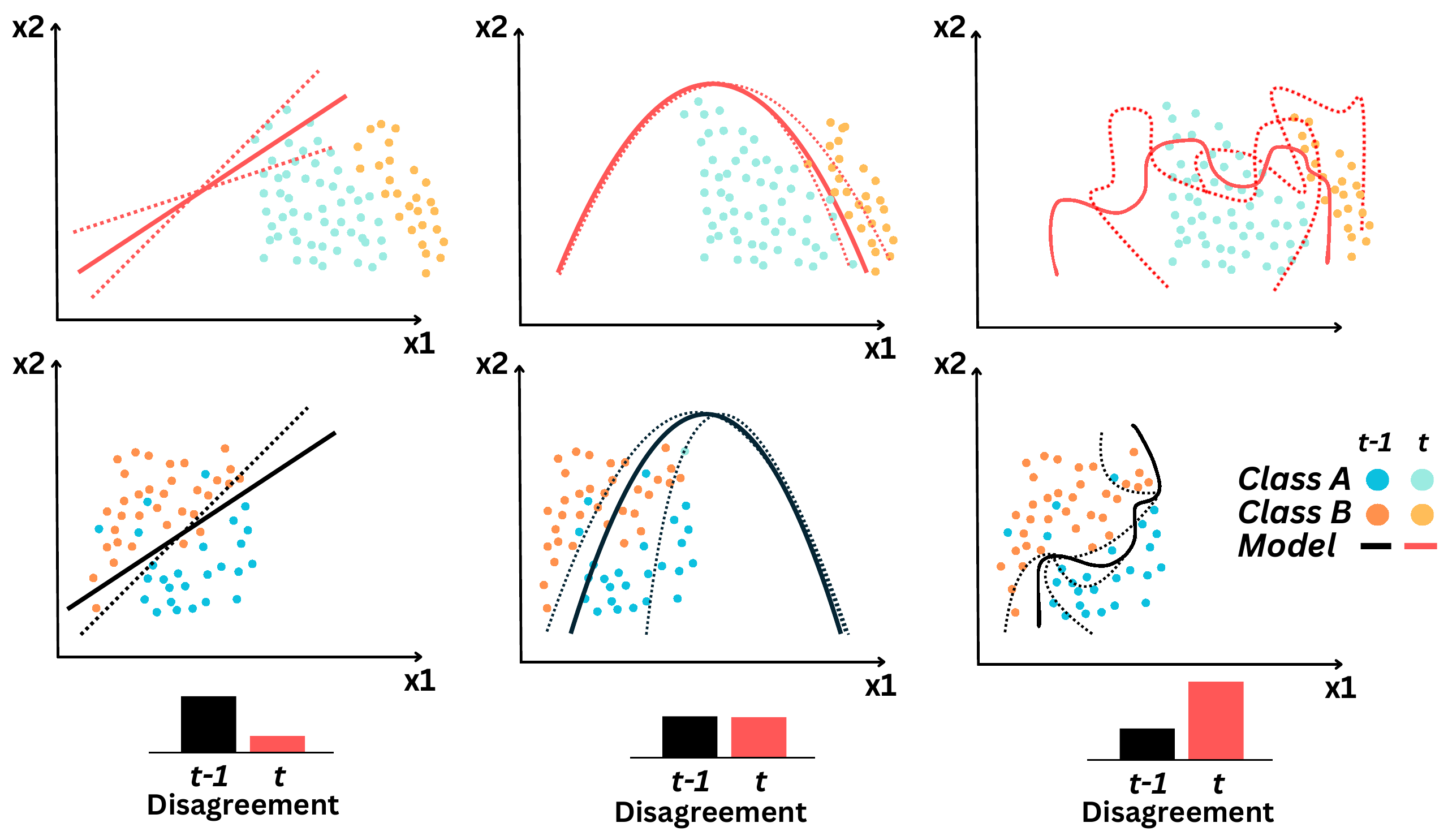}
    \vspace{-0.5cm}
    \caption{Disagreement-based drift across complexities: (left) hardly induced in far input space, (center) even, (right) easily induced in under-regularized far input space.}
    \label{fig:2}
    \vspace{-0.3cm}
\end{wrapfigure}

While $h^*$ is intractable, it motivates maximizing $\Delta(h_{\theta_{t-1}},h')$ to identify parts of the input space most affected by drift. In binary ensembles, this can be as simple as flipping labels \citep{rosenfeld2023almost,ginsberg2022learning} (Fig.~\ref{fig:2}): under-regularized models fail to capture drift, correctly regularized models balance disagreement, and overly complex models overfit new regions. Using this discrepancy while preserving $\varepsilon_{\mathcal{D}_{t-1}}(h_{\theta_{t-1}})$ allows functional regularization, while graceful forgetting and pruning outdated nodes improve adaptability and free capacity.

\section{Method}

\begin{wrapfigure}{r}{0.5\textwidth}
\vspace{-1cm}
\centering
\begin{minipage}{\linewidth}
\begin{algorithm}[H]
\caption{Disagreement framework}
Initialize ensemble $g$ on past data $P$\; 
\While{stream has new batch}{
    $Q', R' \gets$ pseudo-label and flip in $Q$, $R$\; 
    $g_Q, g_R \gets$ copies of $g$\; 
    Train $g_Q$ on $Q'$, $g_R$ on $R'$\; 
    \For{each ensemble $g_X \in \{g_Q, g_R\}$}{
        \For{each model pair $(g_a, g_b)$ in $g_X$}{
            $d_{a,b} = \frac{1}{K} \sum\limits_{i=1}^{K} \mathbf{1}[g_a(x_i) \neq g_b(x_i)]$\;
        }
        $D_X \gets$ collection of all $d_{a,b}$\;
    }
    \If{KS\_test($D_Q, D_R$) rejects $H_0$}{Drift detected\;}
}
\end{algorithm}
\end{minipage}
\vspace{-0.7cm}
\end{wrapfigure}

\begin{wrapfigure}{r}{0.50\textwidth}
    \vspace{-0.5cm}
    \centering
    \includegraphics[width=0.50\textwidth]{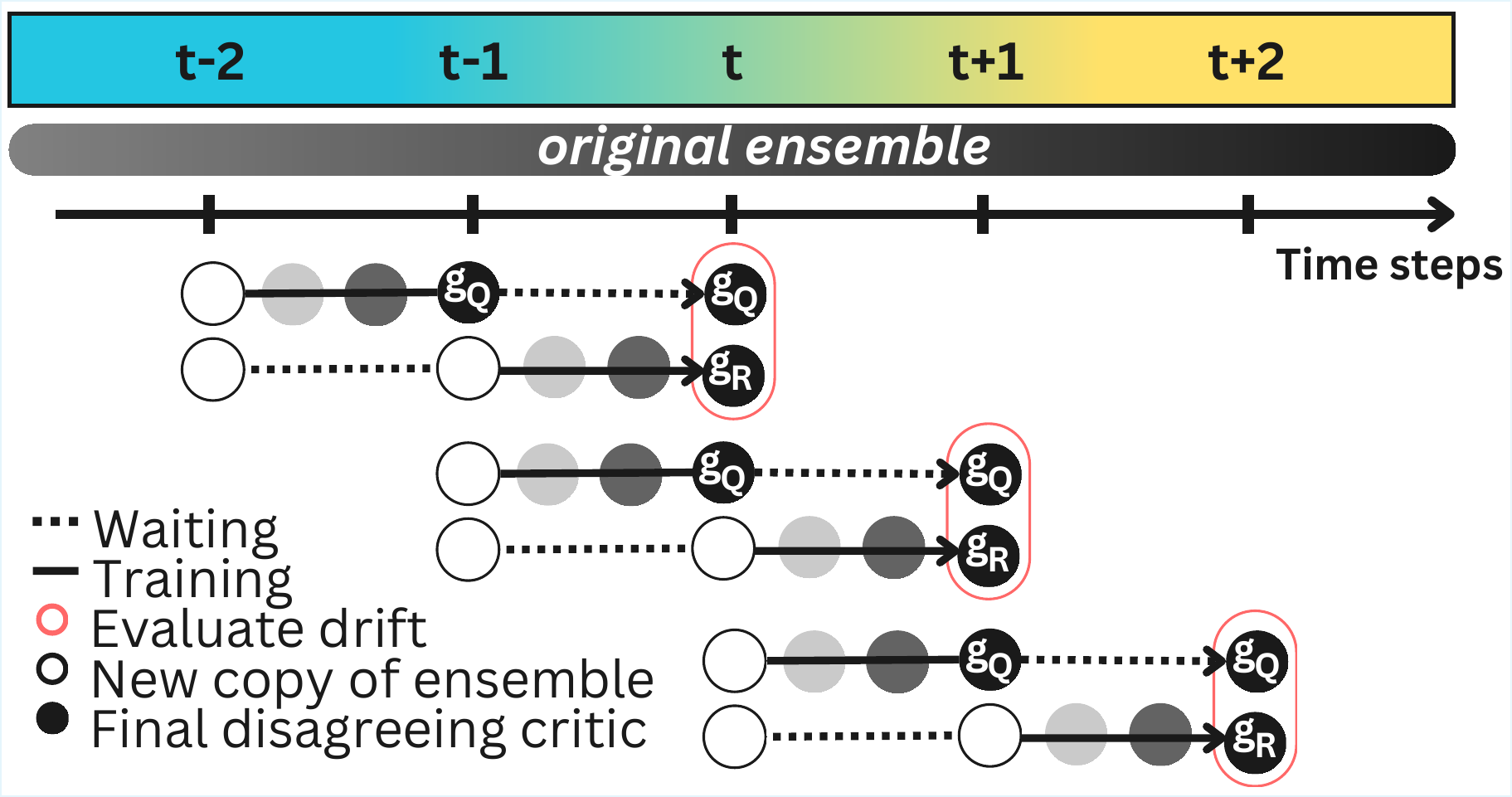} 
    \caption{Windowed disagreement.}
    \label{fig:XXX}
    \vspace{-1.3cm}
\end{wrapfigure}

For each batch, the data is split in two consecutive sub-windows, $Q$ and $R$. Two copies of the ensemble $g$, denoted $g_Q$ and $g_R$, are trained to remain consistent with past distributions $P$ while being exposed to flipped versions of the pseudo-labeled $Q$ and $R$, respectively (Fig. \ref{fig:XXX}). Pairwise disagreements among base learners form the distributions $D_Q$ and $D_R$, capturing the impact of new data on predictive consistency. A Kolmogorov-Smirnov (KS) test between $D_Q$ and $D_R$ is used to detect significant concept drift. The $Q$--$R$ split naturally balances convergence and detection latency, as overly small windows may yield noisy estimates.

To achieve expressive adaptation, without relying on overly large windows that delay detection, we adopt Oza's ensemble backbone, with the Poisson parameter $\lambda$ governing resampling \citep{oza2001experimental}. However, rather than using $\lambda = 1$, instances are exploited more aggressively under underfitting, using $\lambda(\epsilon) = \epsilon \lambda_{\max}$, where $\epsilon \in \langle 0,1 \rangle$ denotes the current error \citep{korycki2022instance}. Thus, accelerating convergence to more reliable estimates.

\section{Experiments}

We evaluate IDT \& MLP ensembles with 6 loss-based: HDDM\textsubscript{A\&W} \citep{pesaranghader2016fast}, ADWIN \citep{adwin}, PH \citep{mouss2004test}, DDM \citep{gama2004learning}, EDDM \citep{baena2006early}; and 5 data-based: BNDM \citep{Xuan2021}, CSDDM \citep{Wan2021}, D3 \citep{sethi2016md3}, IBDD \citep{Souza2020}, OCDD \citep{gozuaccik2021concept}.

\begin{wrapfigure}{r}{0.6\textwidth}
    \vspace{-0.4cm}
    \centering
    \includegraphics[width=0.6\textwidth]{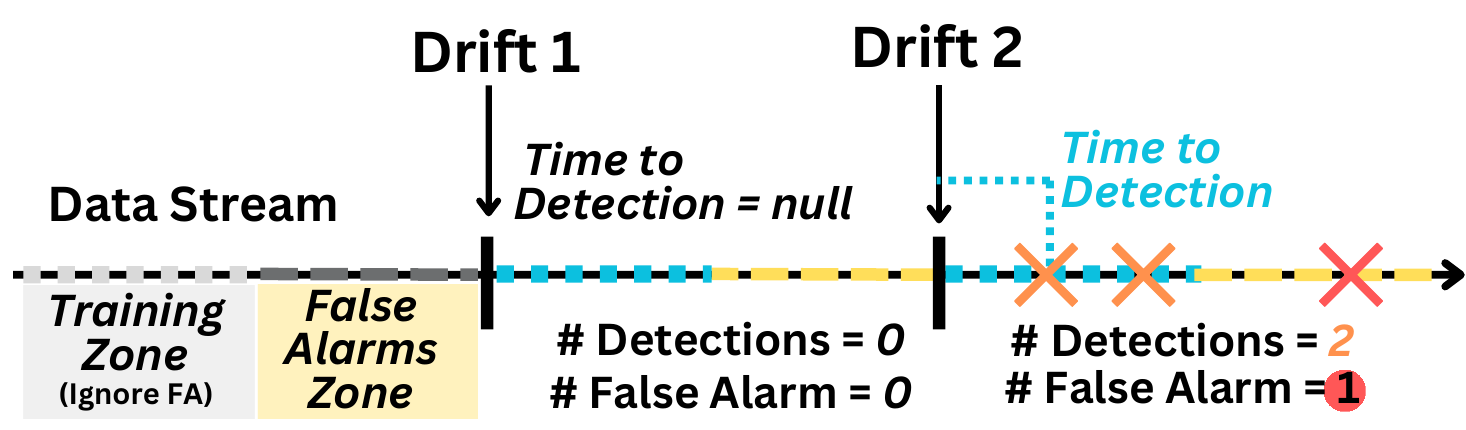} 
    \caption{Evaluation metrics: Detection window.}
    \label{fig:instructions}
    \vspace{-0.2cm}
\end{wrapfigure}

We use 12 synthetic streams from 7 SOA generators: SEA (rotating boundaries), Hyperplane (10 features), Stagger (feature distribution changes), Anomaly Sine (contextual drifts), RBF (centroid shifts), and Agrawal (classification changes). Each contains 90,000 instances with five 15,000-instance drifts, both abrupt and recurring. We adopt prequential evaluation and report Mean Time to Detection (MTD), Detection Accuracy (DA), and False Alarms (FA), counting alarms outside the defined detection window as false positives, with 7,500 and 9,000 instances for abrupt and gradual drifts, respectively (Fig. \ref{fig:instructions}). All hyperparameters for ensembles and drift detectors, including both loss- and data-based methods, were set according to recommended ranges in the original papers and tuned using a weighted min-max normalization:  $0.5 \times \text{DA} + 0.3 \times (1 - \text{FA}) + 0.2 \times (1 - \text{MTD})$. For ensembles, we use as base classifiers: Hoeffding tree \citep{domingos2000mining}, Hoedffing Adaptive Tree \citep{bifet2009adaptive}, and Extremely Fast Decision Tree \citep{manapragada2018extremely} for IDTs, and standard feedforward networks for MLPs, with all ensembles configured to contain 100 learners.

While ensembles of MLPs show good behavior, disagreement-based uncertainty from IDTs performs consistently poorly across nearly all evaluated streams (Table~\ref{tab_combined_drifts}). It exhibits substantially delayed detections and, in several settings, a non-trivial number of false alarms, particularly when compared to loss-based baselines. These results indicate that disagreement signals derived from ensembles of IDTs are often too weak or too noisy to serve as reliable drift indicators.

\begin{table}[h!]
\centering
\setlength{\tabcolsep}{1.5pt} 
\scriptsize
\caption{MTD(FA) results for gradual ($G$) and abrupt ($A$) drifts, in $\otimes$ Disagreement-based, $\diamond$ Data-based, and $\nabla$ Loss-based detectors.}
\label{tab_combined_drifts}
\begin{tabular}{ll c cccccccccc}
\toprule
\textbf{} & \textbf{Method} & \textbf{T} & \textbf{RBF} & \textbf{RBF2} & \textbf{SEA0} & \textbf{SEA1} & \textbf{SEA2} & \textbf{SineA} & \textbf{Sine4} & \textbf{SineL} & \textbf{Hyp0} & \textbf{Hyp1} \\
\midrule
\multirow{4}{*}{$\otimes$} & \multirow{2}{*}{MLPs} & G 
& 1137(4) & 1383(4) & 843(3) & 1475(1) & 2427(1) & 643(3) & 2863(1) & 1747(4) & 1573(2) & 2187(1) \\
 & & A 
& 820(6) & 910(4) & 365(1) & 980(0) & 1620(1) & 410(1) & 1980(0) & 810(0) & 685(1) & 490(0) \\
 \addlinespace
 & \multirow{2}{*}{IDTs} & G & 2267(6) & 1333(15) & 1167(4) & 3700(2) & 3300(3) & 2100(2) & 6600(0) & 3900(3) & 1533(10) & 2467(2) \\
 & & A & 3133(14) & 2840(12) & 2025(0) & 1775(6) & 2275(5) & 1600(13) & 1367(0) & 1200(0) & 1400(17) & 2000(18) \\
\midrule
\multirow{10}{*}{$\diamond$} & \multirow{2}{*}{BNDM} & G & 321(18) & 1825(22) & 1938(11) & 1387(17) & 3221(3) & 566(19) & 180(17) & 180(18) & 2029(9) & 2029(9) \\
 & & A & 152(53) & 89(48) & 173(0) & 186(3) & 175(12) & 65(0) & 133(3) & 132(3) & 1893(40) & 1893(40) \\
 \addlinespace
 & \multirow{2}{*}{CSDDM} & G & 244(36) & 95(52) & 1448(10) & 1245(4) & 240(31) & 801(11) & 851(12) & 381(6) & 137(24) & 108(19) \\
 & & A & 73(80) & 86(96) & 1085(4) & 304(18) & 156(17) & 55(6) & 246(14) & 203(37) & 565(74) & 419(59) \\
 \addlinespace
 & \multirow{2}{*}{D3} & G & 1662(10) & 434(16) & 505(14) & 486(17) & 231(14) & 436(5) & 978(8) & 421(4) & 639(14) & 452(12) \\
 & & A & 123(5) & 121(3) & 728(44) & 583(50) & 547(40) & 129(0) & 129(0) & 129(0) & 1036(53) & 1005(56) \\
 \addlinespace
 & \multirow{2}{*}{IBDD} & G & 92(60) & 139(52) & 254(39) & 379(37) & 168(39) & 343(11) & 158(20) & 101(11) & 232(31) & 232(31) \\
 & & A & 60(154) & 60(147) & 86(113) & 57(100) & 61(90) & 59(18) & 231(3) & 212(0) & 64(108) & 64(108) \\
 \addlinespace
 & \multirow{2}{*}{OCDD} & G & 90(67) & 76(71) & 249(71) & 249(71) & 249(71) & 68(71) & 495(18) & 208(45) & 249(71) & 249(71) \\
 & & A & 189(182) & 188(182) & 149(190) & 149(190) & 149(190) & 62(83) & 138(62) & 138(58) & 153(190) & 153(190) \\
\midrule
\multirow{12}{*}{$\nabla$} & \multirow{2}{*}{ADWIN} & G & 1333(5) & 3733(1) & 4017(1) & 1700(0) & 5117(0) & 2633(1) & 4650(0) & 7325(0) & 2083(2) & 3600(2) \\
 & & A & 1870(7) & 1533(5) & 320(2) & 1680(0) & 1750(2) & 275(0) & 500(0) & 470(0) & 190(6) & 360(3) \\
 \addlinespace
 & \multirow{2}{*}{DDM} & G & 5752(0) & 1336(0) & 4052(0) & 4292(0) & 5160(0) & 585(1) & 4417(0) & 3129(0) & 3091(0) & 3858(0) \\
 & & A & 1839(4) & -(0) & 486(0) & 731(0) & 1321(0) & 317(0) & 280(0) & 269(0) & 336(0) & 1209(0) \\
 \addlinespace
 & \multirow{2}{*}{EDDM} & G & 2029(5) & 4270(0) & 2605(0) & 1994(0) & 7545(0) & 515(2) & -(0) & -(0) & 3063(0) & 3008(0) \\
 & & A & 974(37) & -(0) & 429(8) & 43(0) & 3688(4) & 622(0) & 2026(0) & 2004(0) & 530(29) & 611(8) \\
 \addlinespace
 & \multirow{2}{*}{HDDM\textsubscript{A}} & G & 1037(3) & 3589(3) & 1648(0) & 1604(1) & 1857(1) & 3705(0) & 1144(0) & 3520(1) & 2995(2) & 1710(0) \\
 & & A & 169(4) & 1781(8) & 71(0) & 257(3) & 499(6) & 5(0) & 73(0) & 64(0) & 33(2) & 61(2) \\
 \addlinespace
 & \multirow{2}{*}{HDDM\textsubscript{W}} & G & 1018(3) & 2518(10) & 1840(0) & 1008(3) & 286(8) & 865(0) & 2247(0) & 2284(0) & 3358(0) & 1487(5) \\
 & & A & 1390(19) & 878(4) & 159(0) & 250(2) & 48(8) & 186(2) & 18(5) & 12(7) & 65(2) & 57(1) \\
 \addlinespace
 & \multirow{2}{*}{PH} & G & 1363(7) & 1474(3) & 2060(1) & 1937(0) & 2884(0) & 6466(1) & 2062(0) & 1225(0) & 1434(4) & 1769(3) \\
 & & A & 1364(7) & 1562(6) & 249(1) & 144(0) & 184(1) & 1916(0) & 114(0) & 103(0) & 152(7) & 170(8) \\
\bottomrule
\end{tabular}
\end{table}

\section{Conclusions}

\begin{figure}[h]
    \centering
    \begin{subfigure}[t]{0.23\textwidth}
        \centering
        \includegraphics[width=\textwidth]{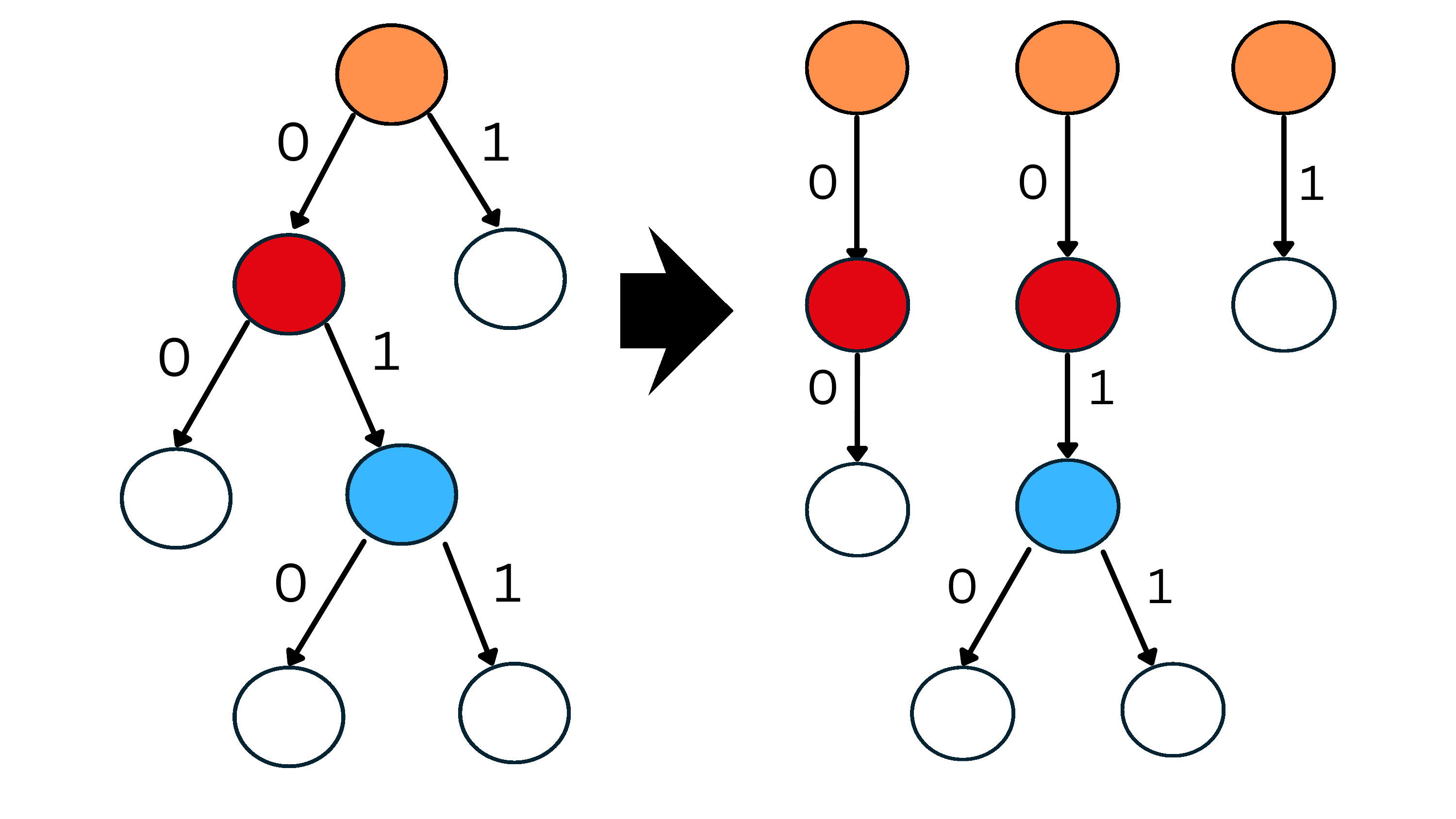}
        \caption{Disconnect subtree}
        \label{fig:step1}
    \end{subfigure}
    \hfill
    \begin{subfigure}[t]{0.26\textwidth}
        \centering
        \includegraphics[width=\textwidth]{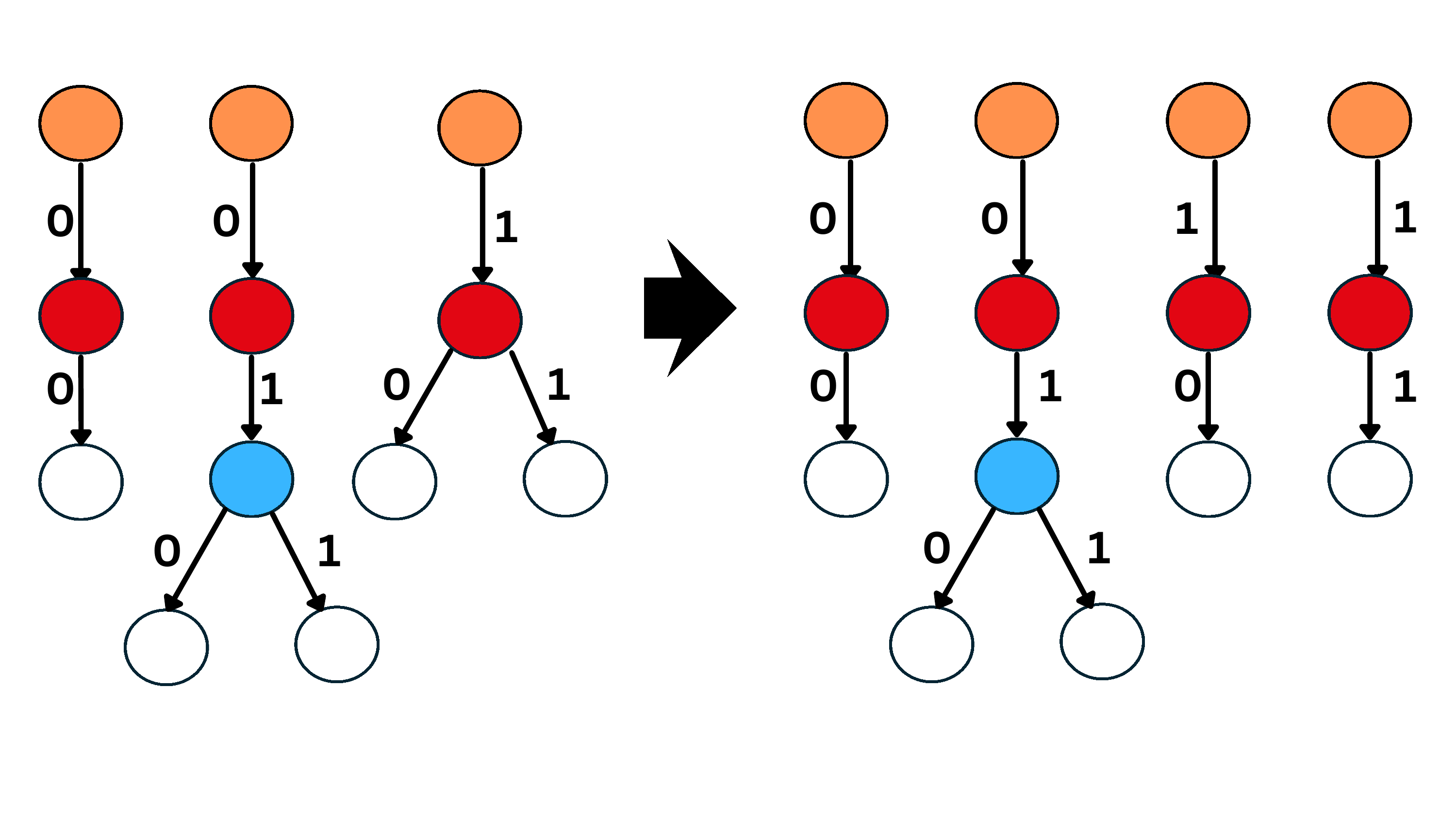}
        \caption{Desired branch splits}
        \label{fig:step2}
    \end{subfigure}
    \hfill
    \begin{subfigure}[t]{0.23\textwidth}
        \centering
        \includegraphics[width=\textwidth]{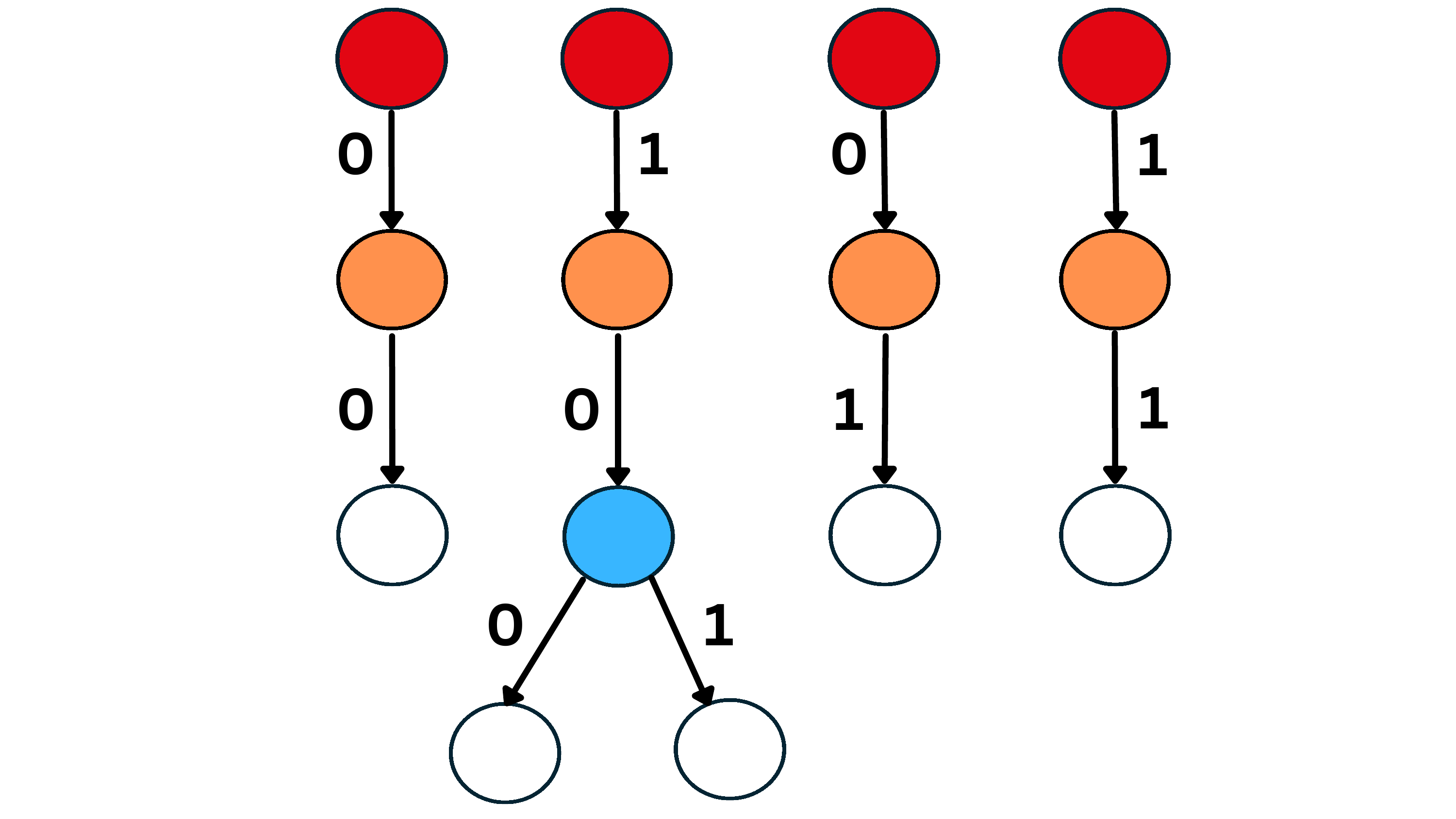}
        \caption{Move splits to root}
        \label{fig:step3}
    \end{subfigure}
    \hfill
    \begin{subfigure}[t]{0.26\textwidth}
        \centering
        \includegraphics[width=\textwidth]{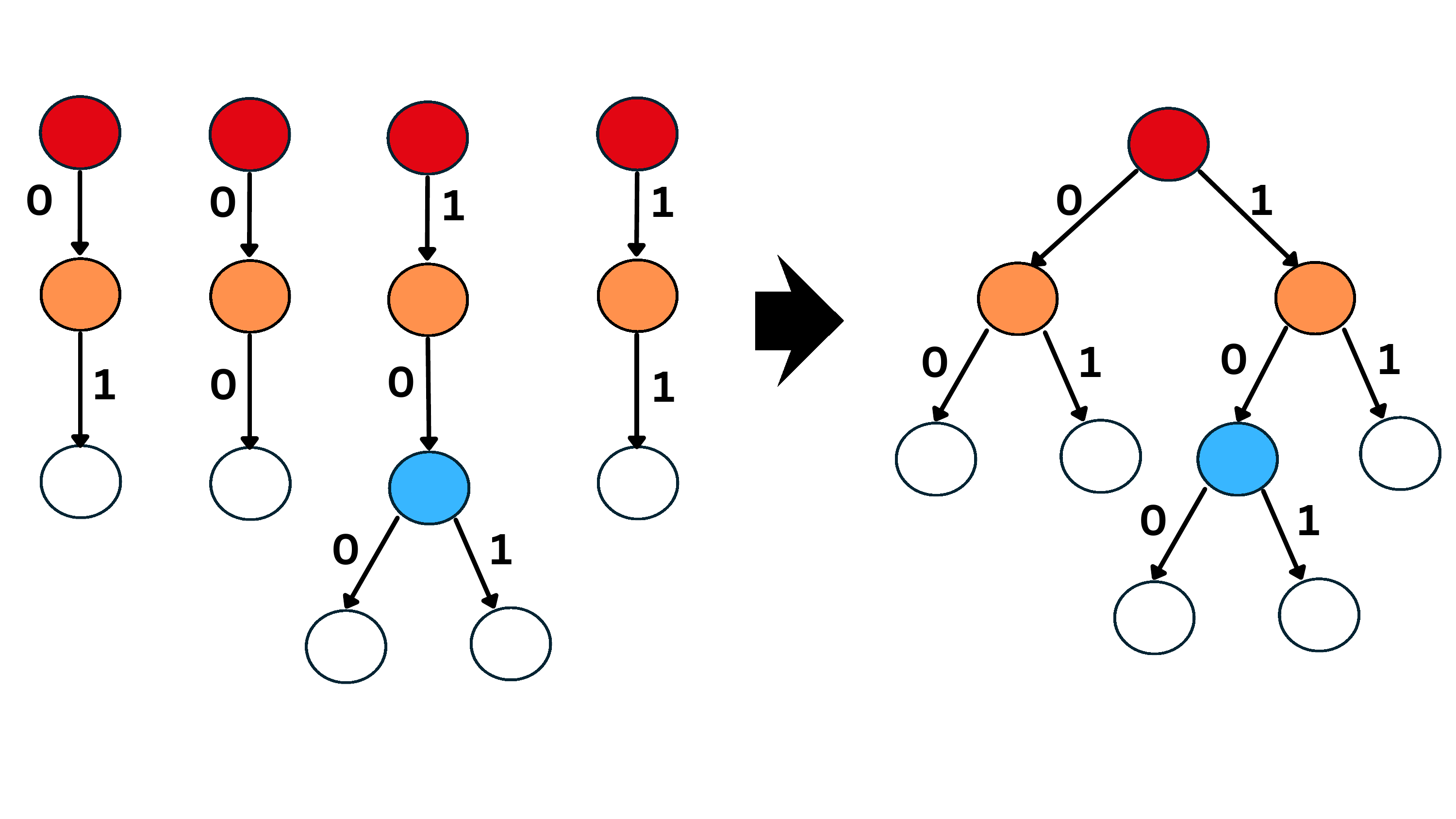}
        \caption{Subtree rebuild}
        \label{fig:step4}
    \end{subfigure}
    \caption{Restructuring IDTs with their intrinsic, non-overlapping rules that fully partition the space.}
    \label{fig:restructuring}
\end{figure}

Taken together, our results hint at a fundamental limitation in current drift research: increasingly sophisticated model-dependent detection mechanisms cannot compensate for rigid base learners. Disagreement estimates derived from IDTs fail to provide reliable signals of concept change, not due to flaws in the detection logic itself, but because the underlying learners lack the plasticity required for uncertainty to reflect learning potential. IDTs converge quickly online thanks to their few trainable parameters, but this efficiency comes at the cost of severe rigidity. Unlike MLP systems, which adapt through both parameter updates and activation dynamics \citep{lourencco2025bridging}, IDTs rely almost exclusively on irreversible structural growth driven by locally optimal split decisions, resulting in history-dependent models dominated by outdated inductive biases \citep{lourencco2025dfdt}. Traditional attempts to address this limitation frame plasticity primarily as capacity management, via subtree pruning \citep{nowak2025behavioral,manapragada2018extremely}, rather than as the ability of the current parameters to serve as a meaningful starting point for further learning. As a consequence, both disagreement-based drift detections exhibits brittle, stream-specific behavior, often failing outside narrow settings. To circumvent this, recent work on restructuring incremental decision trees with their intrinsic, non-overlapping rules \citep{schreckenberger2020restructuring,heyden2024leveraging,zhao2025online} (Fig.~\ref{fig:restructuring}) offer a promising path forward by partially relaxing this rigidity.

\subsubsection*{Acknowledgments}
Work funded by Portuguese Foundation for Science and Technology under the UT Austin Portugal Program, Ph.D. scholarship PRT/BD/18497/2024 and project doi.org/10.54499/UID/00760/2025.

\bibliography{iclr2026_conference}
\bibliographystyle{iclr2026_conference}

\appendix

\end{document}